\documentclass[twocolumn]{article}

\usepackage{arxiv}

\usepackage{microtype}
\usepackage{graphicx}
\usepackage{subfigure}
\usepackage{booktabs} 

\usepackage{hyperref}

\usepackage{amsmath}
\usepackage{amssymb}
\usepackage{mathtools}
\usepackage{amsthm}
\usepackage{enumitem}
\usepackage{wrapfig}
\usepackage{xcolor}
\usepackage{natbib}
\usepackage{authblk}
\usepackage[capitalize,noabbrev]{cleveref}
\usepackage{pifont}

\title{TDMPBC: Self-Imitative Reinforcement Learning for Humanoid Robot Control}

\author{
    Zifeng Zhuang$^{1*}$ \quad
    Diyuan Shi$^{1*}$ \quad
    Runze Suo$^{1}$ \quad
    Xiao He$^{1}$ \quad
    Hongyin Zhang$^{1}$ \protect\\
    Ting Wang$^{1\dag}$ \quad
    Shangke Lyu$^{1\dag}$ \quad
    Donglin Wang$^{1\dag}$ \\
    $^{1}$Westlake University \quad
    $^{*}$Equal Contribution \quad
    $^{\dag}$Corresponding Authors \\
}

\begin{document}

\twocolumn[{
  \begin{@twocolumnfalse}

	\maketitle

	\begin{abstract}
            Complex high-dimensional spaces with high Degree-of-Freedom and complicated action spaces, such as humanoid robots equipped with dexterous hands, pose significant challenges for reinforcement learning (RL) algorithms, which need to wisely balance exploration and exploitation under limited sample budgets.
            In general, feasible regions for accomplishing tasks within complex high-dimensional spaces are exceedingly narrow. 
            For instance, in the context of humanoid robot motion control, the vast majority of space corresponds to falling, while only a minuscule fraction corresponds to standing upright, which is conducive to the completion of downstream tasks.
            Once the robot explores into a potentially task-relevant region, it should place greater emphasis on the data within that region.
            Building on this insight, we propose the \textbf{S}elf-\textbf{I}mitative \textbf{R}einforcement \textbf{L}earning (\textbf{SIRL}) framework, where the RL algorithm also imitates potentially task-relevant trajectories.
            Specifically, trajectory return is utilized to determine its relevance to the task and an additional behavior cloning is adopted whose weight is dynamically adjusted based on the trajectory return.
            As a result, our proposed algorithm achieves 120\% performance improvement on the challenging HumanoidBench with 5\% extra computation overhead. 
            With further visualization, we find the significant performance gain does lead to meaningful behavior improvement that several tasks are solved successfully.
	\end{abstract}

  \end{@twocolumnfalse}
  \vspace{1em}
}]

\section{Introduction}
Humanoid robots with dexterous hands have vast and promising application scenarios due to their behavior flexibility and human-like morphology \citep{bonci2021human, stasse2019overview, choudhury2018humanoid}.
Unfortunately, such complex high-dimensional space is extremely challenging for policy learning with online reinforcement learning (RL) \citep{sferrazza2024humanoidbench, peters2003reinforcement}, which interactively explores the environment and learns optimal decision-making from scratch under the guidance of reward function.
This paradigm has achieved significant success in fields like gaming AI \citep{silver2016mastering, silver2017mastering, schrittwieser2020mastering, hessel2018rainbow} and quadrupedal robots control \citep{miki2022learning, lee2020learning, hwangbo2019learning}.
But when facing humanoid robots equipped with dexterous hands, existing RL methods struggle to learn effectively and efficiently. 
Even sample-efficient model-based state-of-the-art (SOTA) algorithms, such as TD-MPC2 \citep{hansen2023td} and DreamerV3 \citep{hafner2023mastering}, perform poorly in humanoid control.

\begin{figure}[t]
    \centering
    \vspace{4pt}
    \includegraphics[width=0.98\linewidth]{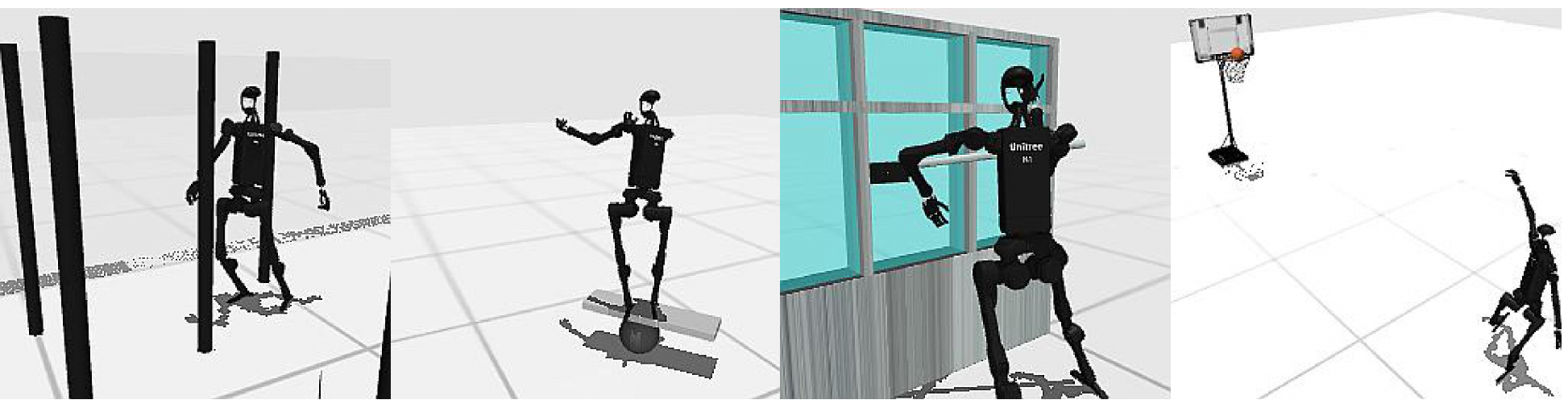}
    \caption{Tasks accomplished by TDMPBC: 1) navigating through pole-filled areas by staying close to the wall, 2) maintain balance on unstable board with the spherical pivot beneath the board in motion, 3) window cleaning with arm-controled cleaning tools and 4) achieving a successful basketball shot.}
    \label{fig:main}
    \vspace{-16pt}
\end{figure}

In high-dimensional complex spaces, the regions capable of accomplishing tasks are typically exceedingly narrow and difficult to explore compared to the entire space. 
For humanoid robot motion control, upright posture is a prerequisite to complete any downstream tasks.
Maintaining an upright posture is similar to balancing an inverted pendulum, where only an extremely small vertical region within the entire space can sustain this posture, while other regions lead to rapid falls.
Therefore, when the algorithm explores an upright posture, the humanoid robot should place particular emphasis on it.

Furthermore, upright posture can be intuitively reflected in return. 
Only if the current timestep maintains an upright posture is it possible to continue obtaining rewards in the following timesteps. 
If the current step results in a fall, given that humanoid robots are virtually incapable of standing up after falling, subsequent rewards become unattainable. 
Under the cumulative effect, the ability to maintain an upright posture will ultimately be reflected very prominently in the return. 
To summarize, the return can be approximated as an indicator of whether the humanoid robot has entered a task-completing region in its control.

Based on the above observation and analysis, we propose a framework called \textbf{S}elf-\textbf{I}mitative \textbf{R}einforcement \textbf{L}earning (SIRL) to assist online learning in complex high-dimensional humanoid robot control.
Building upon the foundation of RL algorithms, SIRL additionally imitates trajectories with high returns.
This enables the humanoid robot to quickly learn the upright posture, thereby accelerating the completion of downstream tasks.
Specifically, we augment the policy training objective in TD-MPC2 \citep{hansen2023td, hansen2022temporal} with an additional behavior cloning term whose weight is dynamically adjusted based on the trajectory return. 
Since the trajectories being imitated are generated by the algorithm itself during exploration, rather than expert demonstrations as in traditional imitation learning (IL), our framework is termed self-imitative. 
Additionally, we refer to TD-MPC2 augmented with a behavior cloning loss as TDMPBC.

\begin{figure}[t]
  \centering
    \includegraphics[width=\linewidth]{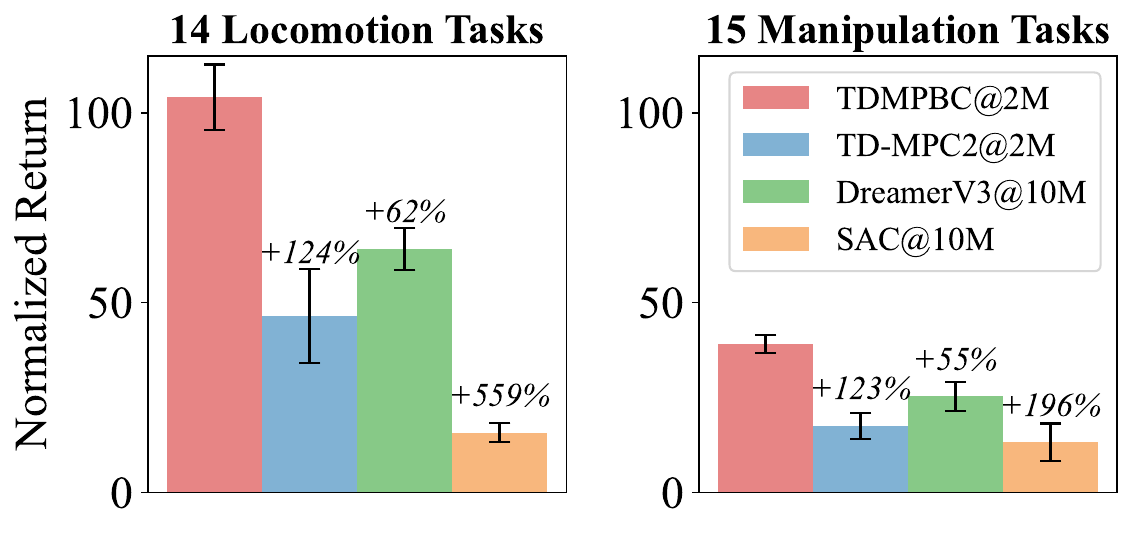}
    \vspace{-16pt}
  \caption{Performance of TDMPBC with 2M interaction steps compared to the baselines TD-MPC2 with 2M, DreamerV3 with 10M and SAC with 10M on HumanoidBench. }
  \label{fig:compare}
  \vspace{-16pt}
\end{figure}

We have validated our proposed TDMPBC on HumanoidBench \citep{sferrazza2024humanoidbench} which contains 31 challenging tasks for the Unitree H1 robot with dexterous hands.
Compared with the baseline TD-MPC2, our proposed method achieves an approximate increase more than 120\% for the normalized return\footnote{We normalize the return to the range $[0, R_{\text{target}}]$. For the \texttt{push} and \texttt{package} manipulation tasks, where the return may be negative, we do not display them in Figure \ref{fig:compare}. The performance of TDMPBC on these two tasks is on par with the baseline, which does not affect the above conclusions we have drawn.}.
What's more, our method enjoys a significantly faster convergence rate and excels in terms of sample-efficiency.
In humanoid locomotion tasks, our algorithm is capable of completing 8 tasks out of 14 with only 2M training steps, whereas the baseline could only accomplish 1 task.

\section{Preliminaries}
\paragraph{Reinforcement Learning} Reinforcement Learning (RL) is a framework of sequential decision.
Typically, this problem is formulated by a Markov Decision Process (MDP) $\mathcal{M}=\{\mathcal{S},\mathcal{A},r,p,d_0,\gamma\}$, with state space $\mathcal{S}$, action space $\mathcal{A}$, scalar reward function $r$, transition dynamics $p$, initial state distribution $d_0(\mathbf{s}_0)$ and discount factor $\gamma$ \citep{sutton1998introduction}.
The objective of RL is to learn a policy $\pi\left(\mathbf{a}_t|\mathbf{s}_t\right)$ at timestep $t$, where $\mathbf{a}_t \in \mathcal{A}$ and $\mathbf{s}_t \in \mathcal{S}$.
Given this definition, the distribution of trajectory $\tau=\left(\mathbf{s}_0, \mathbf{a}_0, \cdots, \mathbf{s}_H, \mathbf{a}_H\right)$ generated by the interaction with the environment $\mathcal{M}$ is $P_{\pi}\left(\tau\right) = d_0(\mathbf{s}_0) \prod_{t=0}^{T} \pi\left(\mathbf{a}_t|\mathbf{s}_t\right) p\left(\mathbf{s}_{t+1}|\mathbf{s}_t,\mathbf{a}_t\right)$,
where $T$ is the length of the trajectory and can be infinite. Then, the goal of RL can be written as an expectation under the trajectory distribution $J\left(\pi\right) = \mathbb{E}_{\tau \sim P_{\pi}\left(\tau\right)}\left[ \sum_{t=0}^{T} \gamma^t r(\mathbf{s}_t, \mathbf{a}_t)\right]$. 
This objective can also be measured by a value function $Q_{\pi}\left(\mathbf{s},\mathbf{a}\right)$, the expected discounted return given the action $\mathbf{a}$ in state $\mathbf{s}$: $Q_{\pi}\left(\mathbf{s},\mathbf{a}\right) = \mathbb{E}_{\tau \sim P_{\pi}\left(\tau|\mathbf{s},\mathbf{a}\right)}\left[ \sum_{t=0}^{T} \gamma^t r(\mathbf{s}_t,\mathbf{a}_t)|\mathbf{s}_0=\mathbf{s},\mathbf{a}_0=\mathbf{a}\right]$. 

\paragraph{TD-MPC2} Model-based RL algorithm TD-MPC2 learns a latent decoder-free world model and selects actions during inference via planning with learned model \citep{hansen2023td}. 
Specifically, TD-MPC2 consists of five components:
\begin{align}
    \nonumber\text{State Encoder:}\qquad\qquad\;\:\mathbf{z}_t &= h_{\phi}\left(\mathbf{s}_t\right), \\
    \nonumber\text{Latent Dynamics:}\qquad\:\mathbf{z}_{t+1} &= d_{\phi}\left(\mathbf{z}_t, \mathbf{a}_t\right), \\
    \nonumber\text{Reward Function:}\qquad\quad \;\hat{r}_t &= R_{\phi}\left(\mathbf{z}_t, \mathbf{a}_t\right), \\
    \nonumber\text{Value Function:}\qquad\quad\enspace\:\: \,\hat{q}_t &= Q_{\phi}\left(\mathbf{z}_t, \mathbf{a}_t\right), \\
    \nonumber\text{Policy Prior:}\qquad\qquad\quad\:\hat{\mathbf{a}}_t &\sim \pi_{\theta}\left(\cdot|\mathbf{z}_t\right),
\end{align}
where $\mathbf{s}_t$ is the states, $\mathbf{a}_t$ is the actions and $\mathbf{z}_t$ is the latent representation. The encoder $h_{\phi}$, dynamics $d_{\phi}$, reward $R_{\phi}$, value $Q_{\phi}$ compose the world model in TD-MPC2 that is trained by minimizing the following objective: 
\begin{align}
    \nonumber \mathcal{L}\left(\phi\right) &= \mathbb{E}_{\left(\mathbf{s}_t, \mathbf{a}_t, r_t, \mathbf{s}_{t+1}\right)_{t=0}^{H} \sim \mathcal{B}}
    \Bigg[\sum_{t=0}^{H}\lambda^t\Big(l^{\left(t\right)}_d+l^{\left(t\right)}_r+l^{\left(t\right)}_q\Big)\Bigg], \\ 
    \nonumber l^{\left(t\right)}_d &= \left\|d_{\phi}\left(\mathbf{z}_t, \mathbf{a}_t\right) - \bar{\mathbf{z}}_{t+1}\right\|_2^2,  \\ 
    \nonumber l^{\left(t\right)}_r &= \text{CE}\Big[R_{\phi}\left(\mathbf{z}_t, \mathbf{a}_t\right) - r_t\Big], \\ 
    \nonumber l^{\left(t\right)}_q &= \text{CE}\bigg[Q_{\phi}\left(\mathbf{z}_t, \mathbf{a}_t\right)-\Big(r_t + \gamma Q_{\bar{\phi}}\big(\mathbf{z}_{t+1}, \pi_{\theta}\left(\cdot|\mathbf{z}_{t+1}\right)\big) \Big)\bigg],
\end{align}
where $\left(\mathbf{s}_t, \mathbf{a}_t, r_t, \mathbf{s}_{t+1}\right)_{t=0}^{H}$ is a trajectory with length $H$ sampled from the replay buffer $\mathcal{B}$, 
$\bar{\mathbf{z}}_{t+1} = h_{\bar{\phi}}\left(\mathbf{s}_{t+1}\right)$ is the target latent representation and CE is the cross-entropy loss. The policy prior $\pi$ is a stochastic maximum entropy policy that learns to maximize the objective: 
\begin{align}
    \nonumber \mathcal{L}\left(\theta\right) = \mathbb{E}_{\mathbf{s}_{t=0}^{H} \sim \mathcal{B}} \Bigg[\sum_{t=0}^{H}\lambda^t\Big[Q_{\phi}\left(\mathbf{z}_t, \pi_{\theta}\left(\cdot|\mathbf{z}_t\right)\right) - \alpha \mathcal{H}\left(\pi\left(\cdot|\mathbf{z}_t\right)\right)\Big]\Bigg],
\end{align}
where $\mathcal{H}$ is the entropy of policy $\pi$ and the parameter $\alpha$ can be automatically adjusted based on an entropy target \citep{haarnoja2018soft} or moving statistics \citep{hafner2023mastering}.

During inference, TD-MPC2 plan actions using a sampling-based planner Model Predictive Path Integral (MPPI) \citep{williams2015model} to iteratively fits a time-dependent multivariate Gaussian with diagonal covariance over the trajectory space such that the estimated return $\hat{\mathcal{R}}$ is maximized:
\begin{align}
    \hat{\mathcal{R}} = \sum_{t=0}^{H-1}\gamma^t R_{\phi}\left(\mathbf{z}_t, \mathbf{a}_t\right) + \gamma^H Q_{\phi}\left(\mathbf{z}_t, \mathbf{a}_t\right).
\end{align}
To accelerate this planning, a fraction of trajectories are generated by the learned policy prior $\pi_{\theta}$.

\paragraph{Imitation Learning} In Imitation Learning (IL) \citep{zare2024survey}, the ground truth reward is not observed and only a set of demonstrations $\mathcal{D}=\left\{\left(\mathbf{s}_t,\mathbf{a}_t\right)\right\}$ collected by the expert policy is provided. The goal of IL is to recover a policy that matches the expert. Behavior cloning \citep{pomerleau1988alvinn} is the most simple and straightforward approach
\begin{align}
    \nonumber\pi_{\text{BC}} = \underset{\pi}{\operatorname{argmax}} \ \mathbb{E}_{\left(\mathbf{s}_t,\mathbf{a}_t\right) \sim \mathcal{D}}\left[\operatorname{log}\pi\left(\mathbf{a}_t|\mathbf{s}_t\right)\right].
\end{align}

\section{Self-Imitative Reinforcement Learning}
In this section, we first highlight the relationship between the motion control of humanoid robots and the upright posture. 
We further observe that maintaining an upright posture corresponds to higher returns. 
Building on this insight, we introduce \textbf{S}elf-\textbf{I}mitative \textbf{R}einforcement \textbf{L}earning (\textbf{SIRL}) and present the implementation based on the model-based algorithm TD-MPC2 \citep{hansen2023td}. 
During the online reinforcement learning process, \textbf{SIRL} provides additional guidance to the humanoid robot to imitate trajectories with high returns. 
Finally, we analyze the characteristics and applicability of this framework from multiple perspectives.

\subsection{Motivation and Analysis}

In humanoid robot motion control, stable upright posture is the essential foundation for both locomotion tasks and whole-body manipulation tasks. 
This point is not only intuitive but also can be reflected in the design of the reward function. 
For example, in Isaac Lab, the reward term related to maintaining upright posture is assigned one extremely high weight:
\begin{align}
    r_1 = 200 \times r_{\text{upright}} + 1.0 \times r_{\text{velocity}} + \cdots,
\end{align}
where $r_{\text{upright}}$ represents an abstraction of all the reward terms associated with maintaining an upright posture, rather than a specific reward term. 
While the term $r_{\text{velocity}}$ rewards velocity tracking.
When tracking different desired velocities, such as 0 m/s, 1 m/s and 5m/s, each corresponds to distinct tasks, namely \texttt{stand}, \texttt{walk}, and \texttt{run}.

Other reward terms are omitted for simplicity. 
In other environments, such as HumanoidBench \citep{sferrazza2024humanoidbench}, 
the upright term $r_{\text{upright}}$ serves as a weight to affect the value of the entire reward function: 
\begin{align}
   r_2 = r_{\text{upright}} \times \left(r_{\text{velocity}} + \cdots\right).
\end{align}
Here $r_{\text{upright}} \in [0, 1]$. 
Intuitively, the upright state with $r_{\text{upright}}=1$ should only occupy an extremely narrow region in the whole state-action space. 
In contrast, the vast majority of the space belongs to $r_{\text{upright}}=0$.
Therefore, once the algorithm explores into the region where an upright posture can be maintained, it should place particular emphasis on it. 
After all, maintaining an upright posture facilitates further exploration and learning for downstream tasks. 
The next question is how to determine whether the upright posture has been explored in the control tasks of humanoid robots.

In reality, for humanoid robots, only standing upright $r_{\text{upright}}=1$ and falling down $r_{\text{upright}}=0$ are stable states. 
Other postures, such as $r_{\text{upright}}=0.6$, will eventually evolve into the stable state of falling. 
Moreover, once a humanoid robot falls, it cannot recover to a standing state on its own, which means that subsequent rewards become unattainable. 
Under the cumulative effect, the ability to maintain a stable standing posture will ultimately be very prominently reflected in the return. We further illustrate this phenomenon with the following example.
\begin{figure}[h]
    \centering
    \includegraphics[width=0.9\linewidth]{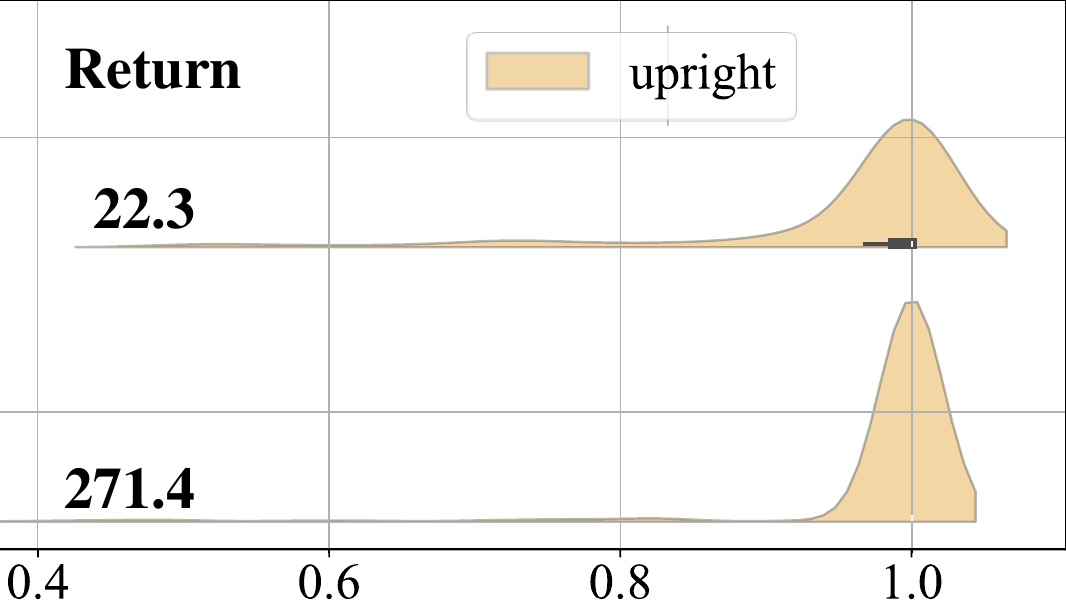}
    \caption{The first row presents the return mean of the trajectories obtained by evaluating the policy trained for 100000 steps on the task \texttt{run}, along with the violin plot distribution of $r_{\text{upright}}$ across all timesteps. The second row shows the results obtained after training for 300000 steps.}
    \label{fig:vio}
\end{figure}

From the Figure \ref{fig:vio}, we observe that the distribution of $r_{\text{upright}}$ obtained from policies trained for 100000 and 300000 steps exhibit relatively small differences. 
But the final returns show a significant disparity (22.3 v.s. 271.4). 
This discrepancy arises because minor differences in whether the humanoid remains upright at each timestep are cumulatively amplified over multiple timesteps.

Overall, the motion control of humanoid robots is closely related to maintaining an upright posture, and whether or not the robot remains upright can lead to a substantial difference in final return.
Based on this finding, an idea for accelerating or assisting humanoid robots emerges: during the online reinforcement learning (RL) exploration process, we can provide additional guidance to the robot to imitate trajectories with high returns. 
This approach enables the robot to first learn how to maintain an upright posture, which then serves as a foundation for completing the entire task.

\subsection{Framework and Methods}
\label{impl_details}
Now we introduce the concept of Self-Imitative Reinforcement Learning (SIRL) which aims to accelerate the learning process of humanoid robots.
During the online exploration process, in addition to the basic RL loss, the policy $\pi_{\theta}$ is also required to imitate trajectories with high returns stored in the replay buffer. 
Unlike classical imitation learning, where trajectories are typically provided by an expert, the trajectories imitated here are generated by the policy \textbf{itself}. 
That is why we call the framework as ``self-imitative''.

We have implemented the SIRL framework based on the TD-MPC2 algorithm \citep{hansen2023td}. 
Specifically, only the policy training loss function is modified and the difference is highlighted by \textcolor{red}{red}:
\begin{align}\label{eq:loss}
    \nonumber \mathcal{L}_{\pi}\left(\theta\right) = &\mathbb{E}_{\left(\mathbf{s}_t, \mathbf{a}_t, \textcolor{red}{R_t}\right)_{t=0}^{H} \sim \mathcal{B}} \Bigg[\sum_{t=0}^{H}\lambda^t\Big[
    \textcolor{red}{\underbrace{\omega\left(R_t\right)\cdot\log \pi_{\theta}\left(\mathbf{a}_t|\mathbf{z}_t\right)
    }_{\text{Self-Imitative}}}
     \Big.\Bigg.\\
    &+\Big.\Bigg. \underbrace{Q_{\phi}\left(\mathbf{z}_t, \pi_{\theta}\left(\cdot|\mathbf{z}_t\right)\right) - \alpha\mathcal{H}\left(\pi\left(\cdot|\mathbf{z}_t\right)\right)}_{\text{Reinforcement Learning}}\Big]\Bigg].
\end{align}
Here $R_t=\sum_{t^{\prime}=0}^{T} \gamma^t r(\mathbf{s}_t, \mathbf{a}_t)$ is the return of the 
whole trajectory and all the $\mathbf{s}_t, \mathbf{a}_t$ within this trajectory have the same return $R_t$.
Compared to the original RL loss function, Another behavior cloning loss is introduced with the weight
\begin{align}
    \omega\left(R_t\right) = \beta \cdot \exp{\left(\frac{R_t - G}{G}\right)},
\end{align}
where $G$ is a reference return value used to determine the level of the current return $R_t$.
Ideally, $G=R_{\text{target}}$ should be the target return, the standard for determining success or failure. 
However, this approach requires the introduction of additional prior information, which may affect the generality of the algorithm. 
Alternatively, we propose using the maximum return $R_{\text{max}}$ of the trajectories in the current replay buffer as the reference standard. 

\subsection{Discussion and Analysis}
From the implementation perspective, our proposed algorithm can be viewed as TD-MPC2 + BC, which might seem similar to the offline algorithm TD3+BC \citep{fujimoto2021minimalist}. 
However, the scenarios and problems they address are completely different. 
As an offline algorithm \citep{zhuang2023behavior, fujimoto2019off, kumar2020conservative}, TD3 + BC incorporates BC \citep{pomerleau1988alvinn} to prevent out-of-distribution (OOD) state-action pairs that lie beyond the offline dataset. 
In contrast, TDMPBC is a fully online algorithm that integrates imitation learning to accelerate exploration and learning within complex high-dimensional spaces.

Generally speaking, imitation learning \citep{zare2024survey} emphasizes exploitation and is a relatively conservative algorithm, whereas reinforcement learning places greater emphasis on exploration. 
TDMPBC can be regarded as a RL algorithm that incorporates conservatism. 
With the dynamic adjustment of BC weights, TDMPBC can be seen as a process where RL explores the space first, followed by rapid learning through imitation learning. 
The introduction of imitation learning does indeed carry the risk of causing the algorithm to converge to local optima. 
However, in the context of high-dimensional complex spaces such as humanoid robot motion control, converging to a local optimum like the upright posture is highly probable and often beneficial for downstream tasks.

\section{Related Work}
Behavior control for Humanoid robots is a long-standing problem, initially explored with simplified humanoid agent \citep{tunyasuvunakool2020dm_control} and recently with full-size humanoid robot \citep{zhuang2024humanoid,fu2024humanplus} such as Unitree H1.
Humanoid robots are of particular interest to the reinforcement learning community because of the high-dimensional action space \citep{merel2017learning, hansen2022temporal, hansen2023td, hansen2024hierarchical}.
To overcome the challenges of exploration in high-dimensional action spaces, some algorithms learn policies by imitating human behavior \citep{fu2024humanplus} or enhance exploration through massive parallelization \citep{zhuang2024humanoid}.
In contrast, our proposed algorithm attempts to learn from scratch without the aid of massive parallelization \citep{makoviychuk2021isaac}. 
We have extensively evaluated our algorithm on the HumanoidBench \citep{sferrazza2024humanoidbench}, a benchmark built on humanoid robot with dexterous hands \citep{menagerie2022github} that contains not only  14 locomotion tasks but also  17 whole-body manipulation tasks.
In the LocoMujoco \citep{al2023locomujoco}, the H1 robot is not equipped with dexterous hands and only focus on locomotion tasks.

Confronted with tasks involving high-dimensional action spaces, model-based RL algorithms \citep{ha2018recurrent, hansen2022temporal, hafner2023mastering, hafner2019dream} often prove to be more sample-efficient compared to model-free alternatives \citep{haarnoja2018soft, fujimoto2018addressing}. 
However, when it comes to humanoid robots with dexterous hands, even the SOTA model-based algorithms struggle to solve it \citep{sferrazza2024humanoidbench}. 
Our algorithm integrates the concept of imitation learning \citep{liu2023ceil, zhang2024context} with the reinforcement learning framework, introducing a loss term of behavioral cloning \citep{pomerleau1988alvinn}.  It may bear a  resemblance to the offline RL \citep{zhuang2024reinformer, fujimoto2019off} algorithm TD3+BC \citep{fujimoto2021minimalist} but our problem setting is   completely different to theirs.
Additionally, it should be noted that the SIRL framework is fundamentally an online RL paradigm that does not rely on expert data, different from IBRL \citep{hu2023imitation} or MoDem \citep{hansen2022modem}.

\section{Experiments}

\begin{figure*}[htbp]
\vspace{-14pt}
  \centering
    \includegraphics[width=0.96\textwidth]{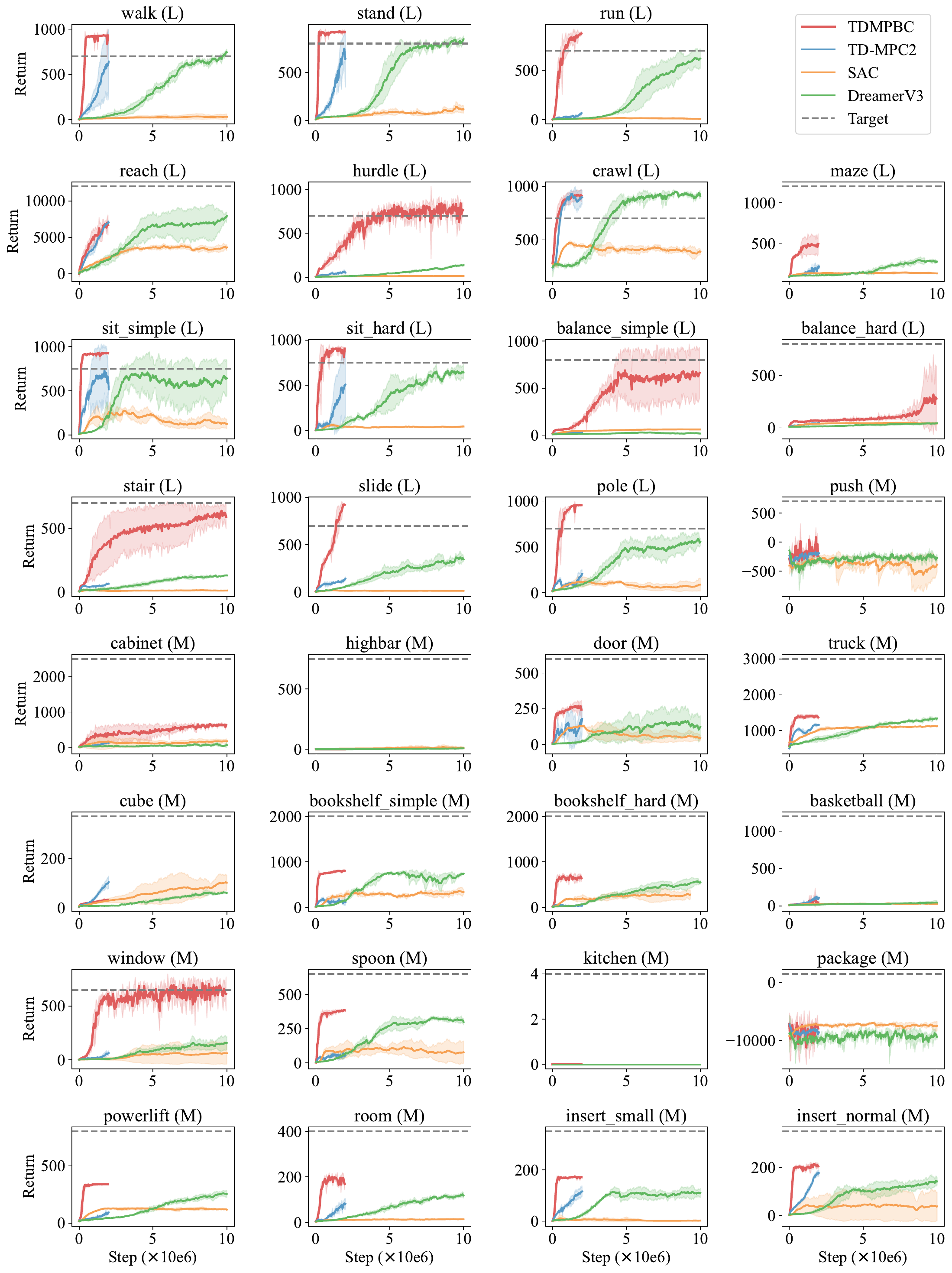}
  \caption{This figure presents the evaluation results on the HumanoidBench, where we conduct experiments with a total of three seeds and the shaded area representing one standard deviation. The baseline results are directly from the HumanoidBench.}\label{fig:main_result}
\end{figure*}
We evaluate our proposed TDMPBC in HumanoidBench \citep{sferrazza2024humanoidbench}, which contains 14 locomotion tasks and 17 whole-body manipulation tasks.
This benchmark is built on the Unitree H1 robot with dexterous hands, which has \textbf{151}-dimension observation space and \textbf{61}-dimension action space.
Remarkably, this benchmark aims to evaluate online RL algorithms and does \textbf{not} include any expert demonstration.
At the same time, SIRL is an improved online RL paradigm rather than imitation learning paradigm.

Specifically, the experiments cover the following five aspects of TDMPBC: 
\textcolor{blue}{1)} Performance comparison with representative RL algorithms across 31 HumanoidBench tasks;
\textcolor{blue}{2)} The impact of the selection of hyperparameter $\beta$ and reference return value $G$;
\textcolor{blue}{3)} Increased runtime caompared to TD-MPC2;
\textcolor{blue}{4)} The phenomenon of policy performance chasing the highest return in the replay buffer during algorithm training;
\textcolor{blue}{5)} Demonstration and analysis of some representative learned behaviors.

\subsection{Performance Comparison}
\subsubsection{Baselines}
We choose these three representative online reinforcement learning algorithms as our baselines:
\begin{itemize}[leftmargin=1.2em]
    \item \textbf{SAC} (Soft Actor-Critic) \citep{haarnoja2018soft}: the state-of-the-art model-free off-policy RL algorithm with maximum entropy learning \citep{eysenbach2021maximum};
    \item \textbf{DreamerV3} \citep{hafner2023mastering}: the state-of-the-art model-based RL algorithm that learns from the imaginary model rollouts;
    \item \textbf{TD-MPC2} \citep{hansen2023td}: the state-of-the-art model-based RL algorithm with online planning achieved via model predictive control (MPC).
\end{itemize}
As for on-policy algorithm PPO (Proximal Policy Optimization) \citep{schulman2017proximal}, its performance is inferior without the massive GPU parallelization so PPO is not our baseline.

In Humanoidbench, TD-MPC2 interacts with the environment for 2M steps, which takes approximately the same amount of time as SAC and DreamerV3 interacting for 10M steps. 
Therefore, the default training steps for TD-MPC2 are set to 2M, while others are set to 10M. 
Similarly, the default training steps for TDMPBC are also 2M.
For tasks where performance significantly surpasses TD-MPC2 but still shows a clear upward trend without reaching the target, we choose to continue training up to 10M steps to demonstrate asymptotic performance. 
These environments include the \texttt{hurdle}, \texttt{balance\_simple}, \texttt{balance\_hard}, and \texttt{stair} tasks in Locomotion, as well as the \texttt{cabinet} and \texttt{window} tasks in Whole-body Manipulation.

\subsubsection{Results on Locomotion} 
HumanoidBench contains a total of 14 locomotion tasks, corresponding to the first 14 training curves ending with (L) in Figure \ref{fig:main_result}. 
It should be noted that, while the locomotion tasks can be accomplished without dexterous hands, the H1 robot here is indeed equipped with dexterous hands. 
The entire robot has \textbf{151}-dimensional observations (51 dimension for body and 50 dimension for each hand), plus a \textbf{61}-dimensional action space, which is quite challenging for RL control.
TDMPBC achieves significantly faster convergence and higher final performance compared to the baseline in all environments except for \texttt{reach} and \texttt{crawl}.
More importantly, TDMPBC surpasses the target (represented by the grey dashed line) in 8 tasks, indicating successful task completion. 
In contrast, the baselines are only capable of completing the \texttt{crawl} task.

\subsubsection{Results on Whole-Body Manipulation} 
HumanoidBench contains a total of 17 whole-body manipulation tasks, corresponding to the last 17 training curves ending with (M) in Figure \ref{fig:main_result}. 
Whole-body manipulation requires not only the control of body posture but also the operation of dexterous hands to accomplish grasping.
Although our algorithm has achieved obvious improvements, the final results are still far behind the target return. 
Our algorithm also struggles to simultaneously control the body and achieve dexterous hand grasping.
A prematurely converging curve implies that the humanoid rapidly masters one thing while the other one fails.

\subsection{Ablation Study}
\subsubsection{Ablation on hyperparameter $\beta$}
\begin{figure}[!b]
    \centering
    \vspace{-28pt}
    \includegraphics[width=0.98\linewidth]{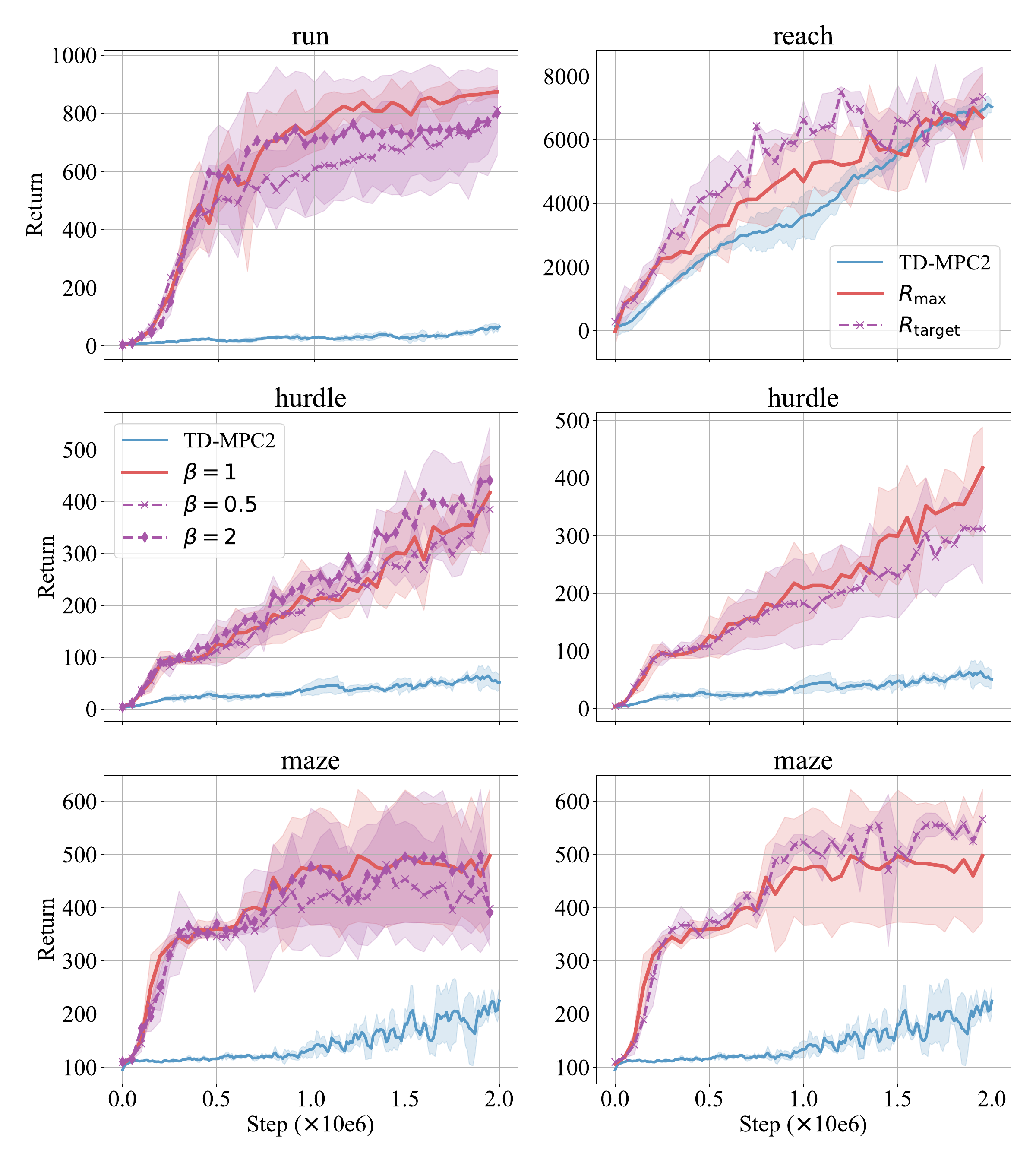}
    \text{~~~~~~(a)~Ablation on $\beta$~~~~~~~(b)~Ablation on $G$}
    \caption{The left figures illustrate the impact of different hyperparameter values ($\beta = 0.5, 1.0, 2.0$) on the performance of TDMPBC across three tasks: \texttt{run}, \texttt{hurdle}, and \texttt{maze}. The right figures demonstrate the effects of two different goal settings ($G = R_{\text{max}}$ and $G = R_{\text{target}}$) on the performance of TDMPBC across three tasks: \texttt{reach}, \texttt{hurdle}, and \texttt{maze}.}
    \label{fig:ablation}
    \vspace{-12pt}
\end{figure}

The $\beta$ balances the original reinforcement learning and our proposed self-imitative behavior cloning in policy loss function Equation \ref{eq:loss}.
Due to the presence of the exponential function and $R_t \leq G$, the range of behavior cloning item is between $(0, 1]$. 
Meanwhile, Q is obtained by discrete regression in a log-transformed space, which means the Q has been normalized \citep{hafner2023mastering}.
Due to the same scale between RL loss and behavior cloning loss, the default value of hyperparameter is $\beta=1$.
In Section 5.1, the experimental results are presented for the case where $\beta=1$. 
To further investigate the robustness of TDMPBC, we conducted additional control experiments with $\beta=0.5$ and $\beta=2.0$. 
We found that the performance of TDMPBC is highly robust to values of $\beta$ around 1.

\begin{figure*}[htbp]
  \centering
  \hspace{-4.5pt}
    \includegraphics[width=0.98\textwidth]{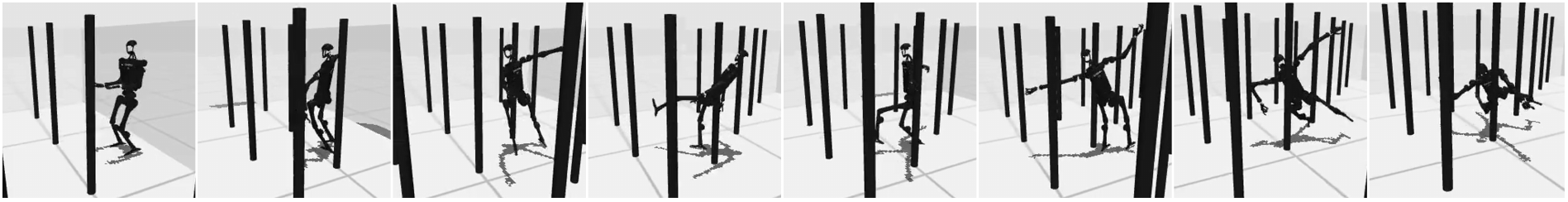}
    \caption{The visualization of baseline TD-MPC2 on the \texttt{pole} task. The robot collides with the pole and falls to the ground.}
    \label{fig:pole_base}
  \hspace{-6pt}
    \includegraphics[width=0.98\textwidth]{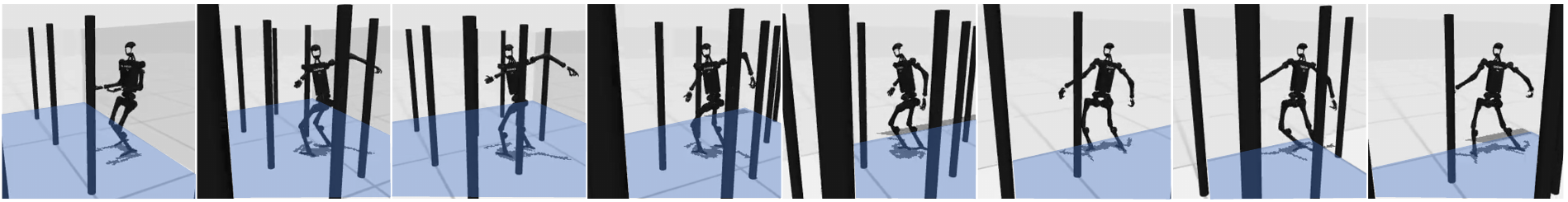}
  \caption{The visualization of our TDMPBC on the \texttt{pole} task. To avoid collisions, the robot chooses to stay close to the wall, thereby passing through quickly and stably. The ground is marked in blue to more clearly illustrate the moving toward the wall.}
  \label{fig:pole_ours}
  \vspace{-8pt}
\end{figure*}
\subsubsection{Ablation on reference return value $G$}
The $G = R_{\text{target}}$ of the current task serves as a globally optimal benchmark but necessitates the introduction of additional information. In contrast, the maximum return $G = R_{\text{max}}$ in the current replay buffer represents the upper limit achievable by the current policy, which grows in tandem with the policy's performance as training progresses.
Meanwhile, $G = R_{\text{target}}$
functions more as a pre-established, relatively higher objective. 
In Figure \ref{fig:ablation}, we found no significant 
\begin{wrapfigure}[10]{r}{0.5\linewidth}
\vspace{-0.5cm}
\centering
\hspace{-8pt}
\includegraphics[width=1.1\linewidth]{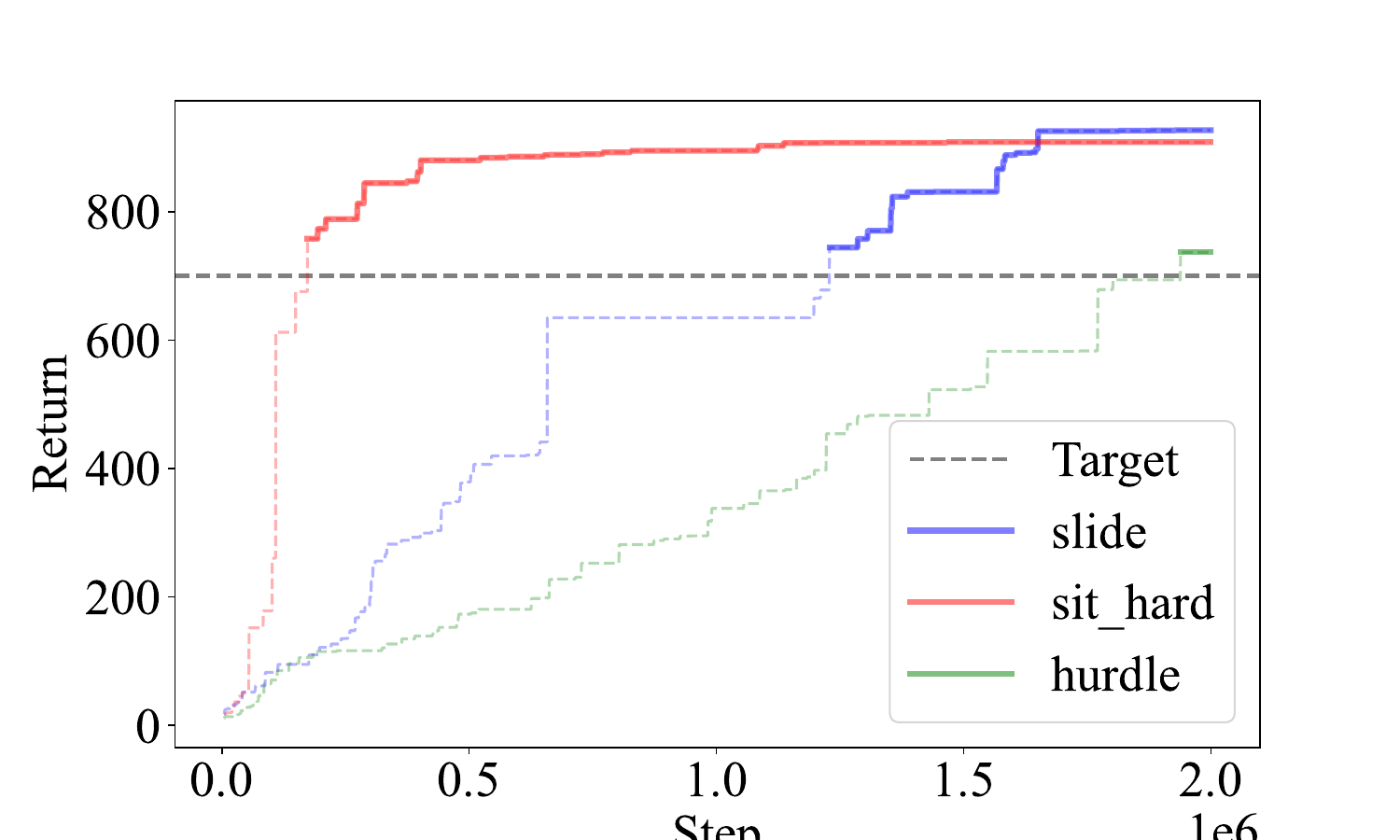}
\vspace{-0.6cm}
\caption{Gradually, $R_{\text{max}}$ may potentially exceed $R_{\text{target}}$.}
    \vspace{-0.4cm}
\end{wrapfigure}
differences in overall performance between the two. 
Therefore, we opted for $G = R_{\text{max}}$ to avoid incorporating prior information. 
It is also worth noting that in tasks where the return can exceed the target, $G = R_{\text{max}}$ may ultimately surpass $G = R_{\text{target}}$.

\subsection{Runtime}
Compared to TD-MPC2, TDMPBC introduces only a marginal increase in computational burden for policy loss calculations and requires the replay buffer to additionally store the current maximum return value. 
We measure the time required to run three seeds simultaneously on a Tesla V100-SXM2-32G GPU across three different tasks in Table \ref{tab:time}. 
When the GPU is upgraded to an NVIDIA A100-SXM4-40GB, the time can be further reduced to approximately 20 hours.
On average, TDMPBC only increases the time by less than $5\%$, yet achieved a remarkable performance improvement of over $120\%$.
\begin{table}[h!]
\centering
\caption{A comparison of the runtime of our proposed algorithm TDMPBC and the baseline TD-MPC2 on the same hardware.}
\label{tab:time}
\vspace{4pt}
\resizebox{\linewidth}{!}{
\begin{tabular}{l|ll|r}
\toprule
Tasks   & TDMPBC          & TD-MPC2          & Improvement \\
\hline
walk    & $37.71\pm0.13$  & $35.23\pm  0.38$ & $+\ 7.06\%$        \\
reach   & $37.73\pm0.64$  & $36.46\pm  0.71$ & $+\ 3.48\%$        \\
hurdle  & $36.61\pm0.06$  & $35.34\pm  0.13$ & $+\ 3.61\%$        \\
\hline
\textbf{\textit{Average}} & $\textbf{\textit{37.35}} $ (h)         & $\textbf{\textit{35.68}}$ (h)         & $+\ \textbf{\textit{4.72\%}}$ \\
\bottomrule
\end{tabular}}
\vspace{-8pt}
\end{table}

\subsection{Training phenomenon}
\begin{figure}[h]
\vspace{-8pt}
    \centering
    \includegraphics[width=0.98\linewidth]{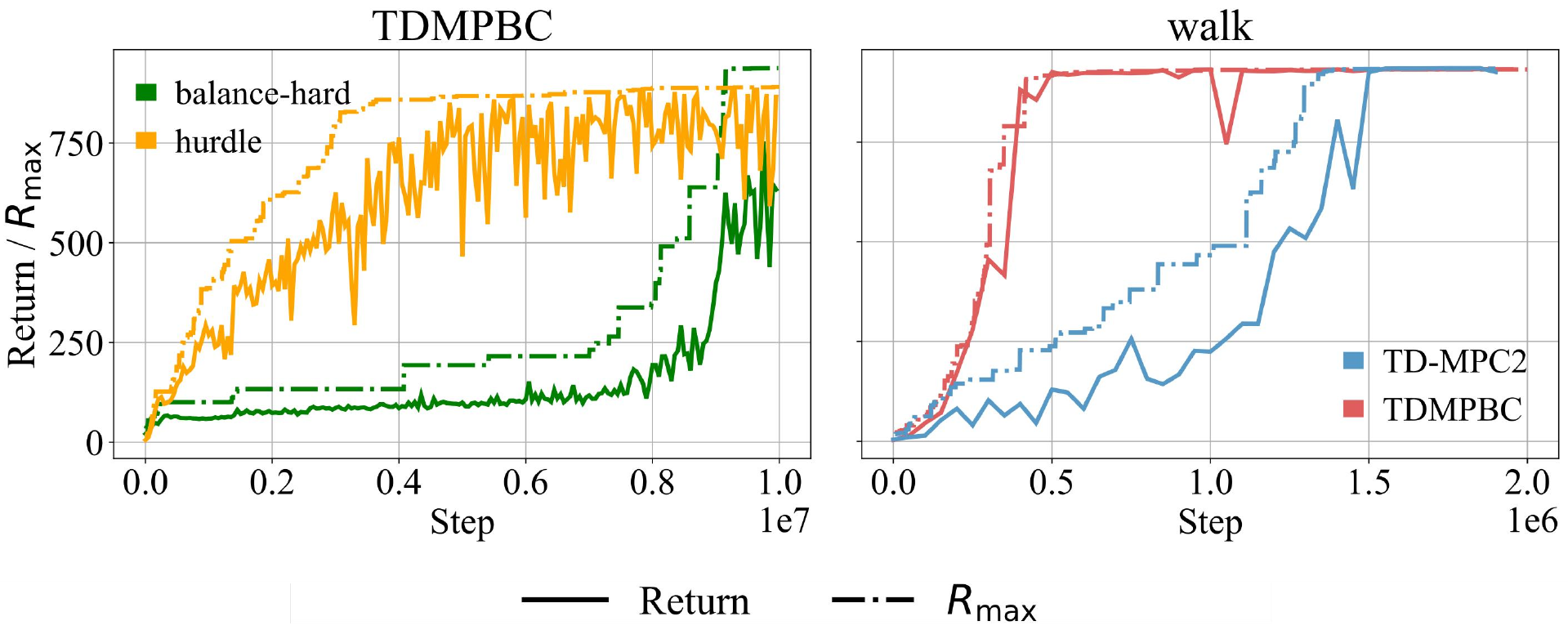}
    \caption{The curves of the policy's return (solid line) and the maximum return in the current replay buffer (dashed line) during training. The left figure shows these curves for TDMPBC on the \texttt{balance-hard} and \texttt{hurdle} tasks, while the right figure compares the curves for TDMPBC and TD-MPC2 on the \texttt{walk}.}
    \label{fig:max_R}
    \vspace{-8pt}
\end{figure}

In the experiments, we observed that the return obtained by evaluating the current policy is often lower than the maximum return in the replay buffer, regardless of whether it is our proposed TDMPBC or the baseline algorithm TD-MPC2 in Figure \ref{fig:max_R}. Intuitively, it appears as though the policy is constantly chasing the current maximum return, and the behavior cloning (BC) in SIRL accelerates this.

\subsection{Behavior Visualization} 
In this subsection, we present representative behaviors obtained from TDMPBC, including the \texttt{pole} and \texttt{balance-hard} tasks in locomotion, as well as the \texttt{window} task in whole-body manipulation.

\textbf{\texttt{pole}}: In the \texttt{pole} task, the humanoid robot is required to navigate forward through a dense forest of tall, slender poles without collision. 
The robot trained with TD-MPC2 continuously collides with the poles and is unable to move forward properly, eventually falling over.
Once step inside the pole forest, the robot swiftly moves towards one of the side walls. 
It then proceeds to hug the wall, keep escaping the dense poles, thereby avoiding potential collisions. 
This clever avoidance strategy, closely resembling human behavior, is exactly the reason our algorithm converges rapidly.

\begin{figure}[h]
    \centering
    \includegraphics[width=0.98\linewidth]{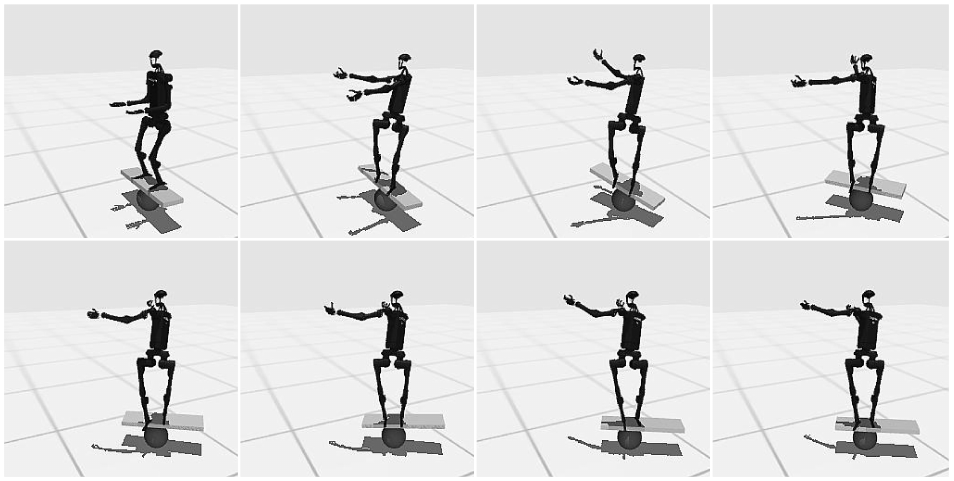}
    \caption{The visualization of our TDMPBC on the \texttt{balance-hard} task, where the humanoid robot aims to maintain balance on an unstable board, with the spherical pivot beneath the board in motion.}
    \label{fig:balance}
\end{figure}
\textbf{\texttt{balance\_hard}}: The humanoid robot is required to maintain balance on an unstable plank, beneath which lies a movable sphere. Upon initialization, the robot nearly falls backward but manages to stabilize itself by swinging its arms, ultimately achieving a balanced posture. During this process, the sphere is displaced from the center of the plank to a position slightly to the left.

\begin{figure}[h]
    \centering
    \includegraphics[width=0.98\linewidth]{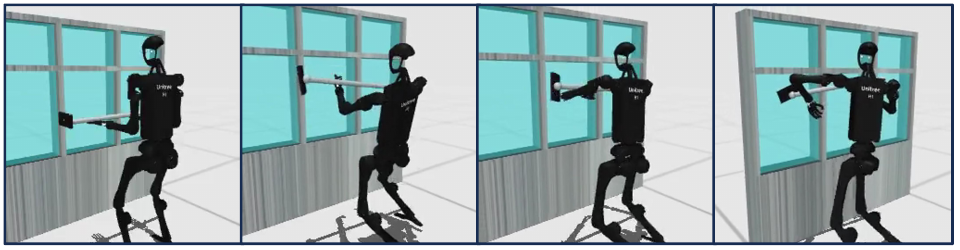}
    \vspace{-6pt}
    \caption{The visualization of our TDMPBC on the \texttt{window} task, where the humanoid aims to grab a window wiping tool and keep its tip parallel to a window by following a prescribed vertical speed.}
    \label{fig:window}
    \vspace{-8pt}
\end{figure}
\paragraph{\texttt{window}}: For the \texttt{window} task, the robot should ideally use its dexterous hands to control the cleaning tools. 
However, the robot does not master the control of its dexterous hands and instead chooses to press against the cleaning surface with its arms to complete the task. 
Despite the algorithm achieving a high return on this task, it does not perform the task as expected.

\section{Conclusion, Limitation and Future Work}
In this paper, we propose the Self-Imitative Reinforcement Learning (SIRL) framework to accelerate the online learning of humanoid robots.
We have made a simple yet effective modification to TD-MPC2 by incorporating a behavior cloning (BC) loss term into the policy training loss function. 
Our proposed algorithm has demonstrated significant performance improvements in the HumanoidBench with a little additional computation overhead. 
Current research has achieved remarkable progress in locomotion tasks, yet there remains substantial room for improvement in whole-body manipulation. 
Looking ahead, we plan to transition TDMPBC from simulated environments to real-world deployment, further exploring its strengths and weaknesses.

\bibliographystyle{unsrtnat}
\bibliography{main}

\newpage
\appendix
\onecolumn

\section{Contributions}
The first version of this paper is completed by the following five authors: \underline{Zifeng Zhuang$^*$}, \underline{Diyuan Shi$^*$}, \underline{Ting Wang\ding{41}}, \underline{Shangke Lyu\ding{41}} and \underline{Donglin Wang\ding{41}}. 
The first version was submitted to ICRA (about \textbf{2024.9}), but it was desk-rejected due to formatting issues. 
\underline{Zifeng Zhuang$^*$} proposes and leads the entire project. 
\underline{Diyuan Shi$^*$} develops the code and conducts all the experiments, and thus is credited as a co-first author. 
Dr. \underline{Ting Wang\ding{41}} implementes the policy evaluation and visualization, participats in the writing of the paper and provides valuable feedback. 
Dr. \underline{Shangke Lyu\ding{41}} provides extensive discussions on article structure, which ensures the smooth progress of the entire project.
Professor \underline{Donglin Wang\ding{41}} fully supports the development of this project, provides all the necessary resources. 

In the second version of the paper (current version), three additional authors contributed including \underline{Runze Suo}, \underline{Xiao He} and \underline{Hongyin Zhang}. 
\underline{Runze Suo} implements the modification of weights and update the code. 
\underline{Xiao He} participats in data preprocessing and visualization of the paper. 
\underline{Hongyin Zhang} offers numerous valuable suggestions for the writing of the paper.
\section{HumanoidBench}
HumanoidBench \citep{sferrazza2024humanoidbench} is a comprehensive simulated humanoid robot benchmark designed to evaluate and advance research in whole-body locomotion and manipulation tasks. It aims to provide a standardized platform for testing and developing algorithms for humanoid robots, addressing the challenges of complex dynamics, sophisticated coordination, and long-horizon tasks.

\paragraph{A.1 Simulated Humanoid Robot:} HumanoidBench features a humanoid robot equipped with two dexterous hands, specifically the Unitree H1 robot with Shadow Hands. 
The robot is simulated using the MuJoCo physics engine, which provides fast and accurate physics simulation.
The environment supports both position control and torque control, with the action space normalized to $\left[-1,1\right]$ for 61 actuators (19 for the body and 21 for each hand).

\paragraph{A.2 Tasks}
HumanoidBench includes a diverse set of 31 tasks, divided into 14 locomotion tasks and 17 whole-body manipulation tasks. These tasks range from simple locomotion (e.g., walking, running) to complex manipulation (e.g., package unloading, tool usage, furniture assembly). Below is the specific list of tasks:

\textbf{Locomotion tasks}
\begin{itemize}
    \item \texttt{Walk}: The robot maintains a velocity of approximately 1 m/s, ensuring it does not fall to the ground.
    \item \texttt{Stand}: The robot maintains a standing pose.
    \item \texttt{Run}: The robot maintains a velocity of approximately 5 m/s, ensuring it does not fall to the ground.
    \item \texttt{Reach}: The robot's left hand reaches a randomly initialized 3D point.
    \item \texttt{Hurdle}: The robot maintains a velocity of approximately 5 m/s while clearing hurdles, ensuring it does not fall to the ground.
    \item \texttt{Crawl}: The robot traverses through a tunnel at a velocity of approximately 1 m/s.
    \item \texttt{Maze}: In a maze, the robot reaches its target location by making multiple turns at intersections.
    \item \texttt{Sit}: In $sit\_simple$, the robot sits on a chair located nearby behind it. The $sit\_hard$ task involves a movable chair, where the robot sits on a chair positioned at random directions and locations.
    \item \texttt{Balance}: The robot maintains its balance on an unstable board. In $balance\_simple$, the spherical pivot beneath the board remains stationary, whereas in $balance\_hard$, the pivot is mobile.
    \item \texttt{Stair}: The robot ascends and descends stairs at a velocity of 1 m/s.
    \item \texttt{Slide}: The robot slides upwards and downwards at a velocity of 1 m/s.
    \item \texttt{Pole}: The robot advances through a dense forest composed of high thin poles, without colliding with them.
\end{itemize}

\textbf{Whole-Body Manipulation tasks}
\begin{itemize}
    \item \texttt{Push}: The robot moves a box to a randomly initialized 3D point on a table.
    \item \texttt{Cabinets}: The robot opens four different types of cabinet doors (such as hinged doors, sliding doors, drawers, and pull-up cabinets) and performs various pick-and-place manipulations inside the cabinets.
    \item \texttt{Highbar}: The robot maintains a grip with both hands on the high bar, swinging while maintaining its hold until it reaches a vertical, upside-down position.
    \item \texttt{Door}: The robot turns the doorknob to open the door, and walks through it while keeping the door open.
    \item \texttt{Truck}: The robot unloads packages from a truck onto a platform.
    \item \texttt{Cube}: The robot manipulates a cube with each hand, aligning both cubes with specific, randomly initialized target orientations.
    \item \texttt{Bookshelf}: The robot rearranges five objects on a bookshelf, which is equivalent to five designated sub-tasks, each involving the positioning of a different object to a target location. The sub-tasks must be completed in order. In $bookshelf\_simple$, the order of sub-tasks is always the same, whereas in $bookshelf\_hard$, the order is randomized.
    \item \texttt{Basketball}: The robot catches a ball coming from a random direction and throws it into the basket.
    \item \texttt{Window}: The robot, holding a window cleaning tool, maintains its tip parallel to the window while adhering to a specified vertical velocity.
    \item \texttt{Spoon}: The robot picks up a spoon located next to a pot and uses it to draw circles in the pot.
    \item \texttt{Kitchen}: The robot performs a series of actions in a kitchen environment, including opening the microwave door, moving a kettle, turning a burner, and switching on and off the lights.
    \item \texttt{Package}: The robot moves a box to a randomly initialized target location.
    \item \texttt{Powerlift}: The robot lifts a barbell of a designed mass.
    \item \texttt{Room}: The robot arranges a 5m by 5m room, minimizing the positional variance in the x and y axes by filling it with randomly dispersed objects.
    \item \texttt{Insert}: The robot inserts the ends of a rectangular block into two small pegs. $Insert\_small$ and $insert\_normal$ indicate different object sizes.
\end{itemize}

\paragraph{A.3 Observation and Action Space} The observation space for the robot state includes joint positions and velocities, totaling 151 dimensions (49 for the body and 51 for each hand). Additionally, the environment provides task-specific observations, such as object positions and velocities, to facilitate interaction with the environment.
The action space is normalized to $[-1,1]$ for 61 actuators, which includes 19 actuators for the humanoid body and 21 actuators for each hand. The action space is designed to be consistent across all tasks to minimize domain-specific tuning.

\begin{figure}[htbp]
  \centering
  \hspace{-6pt}
    \includegraphics[width=0.95\textwidth]{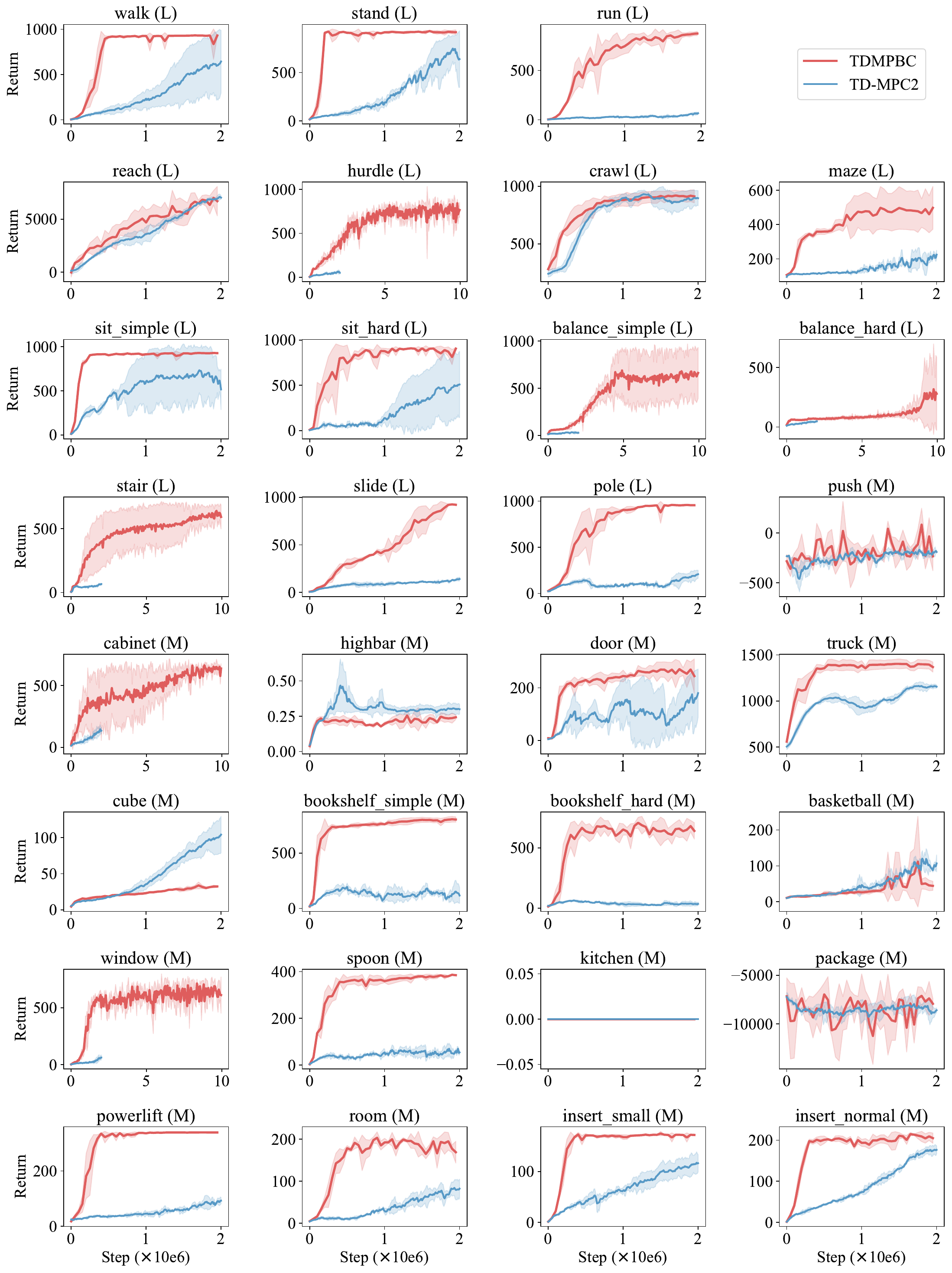}
  \caption{The evaluation curves of our proposed TDMPBC and the baseline TD-MPC2. We conduct experiments with a total of three seeds and the shaded area representing one standard deviation.}
  \label{fig:12}
\end{figure}

\section{Experimental Details}
We implemented our TDMPBC algorithm on the source code of TD-MPC2 in HumanoidBench \url{https://github.com/carlosferrazza/humanoid-bench}. 
The parameters used were the default parameters of HumanoidBench when running TD-MPC2. 
For the additional hyperparameter $\beta$ that we introduced, it was set to $\beta=1$ in all experiments.

\section{More Experiment Results}

More experiment results are presented here. Figure \ref{fig:12}  compares the evaluation curves of our proposed TDMPBC against the baseline TD-MPC2, while Table \ref{tab:results} provides a summary of performance across various methods on HumanoidBench tasks, effectively demonstrating the efficacy of our approach.

\begin{table}[htbp]
\centering
\caption{Summary of Results for HumanoidBench Tasks. In the results, non-target-exceeding scores are displayed in \textcolor{gray}{gray}, and the best results for each task are highlighted in \textbf{bold}.
}
\label{tab:results}
\begin{tabular}{l|l|llll}
\toprule
Tasks & Target & TDMPBC@2M & TD-MPC2@2M & DreamerV3@10M & SAC@10M \\ \hline
\texttt{walk} & 700 & \textbf{932.08 $\pm$ 0.82} & \textcolor{gray}{644.19 $\pm$ 344.25} & 751.02 $\pm$ 28.25 & \textcolor{gray}{36.40 $\pm$ 30.28} \\ 
\texttt{stand} & 800 & \textbf{929.67 $\pm$ 0.69} & \textcolor{gray}{749.79 $\pm$ 133.86} & 845.36 $\pm$ 33.43 & \textcolor{gray}{141.98 $\pm$ 56.35} \\ 
\texttt{run} & 700 & \textbf{874.58 $\pm$ 21.69} & \textcolor{gray}{66.14 $\pm$ 9.97} & \textcolor{gray}{629.33 $\pm$ 81.75} & \textcolor{gray}{18.36 $\pm$ 3.30} \\ 
\texttt{reach} & 12000 & \textcolor{gray}{7013.83 $\pm$ 685.29} & \textcolor{gray}{7120.75 $\pm$ 253.64} & \textbf{\textcolor{gray}{7926.20 $\pm$ 546.66}} & \textcolor{gray}{3800.29 $\pm$ 344.43} \\ 
\texttt{hurdle} & 700 & \textbf{843.54 $\pm$ 63.58} & \textcolor{gray}{64.68 $\pm$ 9.70} & \textcolor{gray}{137.46 $\pm$ 9.07} & \textcolor{gray}{13.85 $\pm$ 8.90} \\ 
\texttt{crawl} & 700 & 920.54 $\pm$ 45.88 & 931.69 $\pm$ 33.19 & \textbf{950.98 $\pm$ 10.38} & \textcolor{gray}{471.95 $\pm$ 12.13} \\ 
\texttt{maze} & 1200 & \textbf{\textcolor{gray}{497.66 $\pm$ 124.50} }& \textcolor{gray}{224.62 $\pm$ 25.65} & \textcolor{gray}{301.77 $\pm$ 36.47} & \textcolor{gray}{149.40 $\pm$ 13.84} \\ 
\texttt{sit\_simple} & 750 & \textbf{928.82 $\pm$ 1.12} & \textcolor{gray}{733.90 $\pm$ 255.79} & \textcolor{gray}{710.96 $\pm$ 208.93} & \textcolor{gray}{275.94 $\pm$ 33.41} \\ 
\texttt{sit\_hard} & 750 & \textbf{908.75 $\pm$ 2.27} & \textcolor{gray}{508.98 $\pm$ 365.77} & \textcolor{gray}{662.55 $\pm$ 22.79} & \textcolor{gray}{61.06 $\pm$ 13.78} \\ 
\texttt{balance\_simple} & 800 & \textbf{\textcolor{gray}{688.86 $\pm$ 239.99}} & \textcolor{gray}{34.07 $\pm$ 4.42} & \textcolor{gray}{29.89 $\pm$ 0.27} & \textcolor{gray}{62.61 $\pm$ 2.73} \\ 
\texttt{balance\_hard} & 800 & \textbf{\textcolor{gray}{317.27 $\pm$ 379.72}} & \textcolor{gray}{48.18 $\pm$ 8.49} & \textcolor{gray}{45.04 $\pm$ 6.13} & \textcolor{gray}{50.82 $\pm$ 2.56} \\ 
\texttt{stair} & 700 & \textbf{\textcolor{gray}{640.37 $\pm$ 46.99}} & \textcolor{gray}{66.50 $\pm$ 6.77} & \textcolor{gray}{132.14 $\pm$ 1.80} & \textcolor{gray}{18.02 $\pm$ 4.91} \\ 
\texttt{slide} & 700 & \textbf{926.26 $\pm$ 2.17} & \textcolor{gray}{141.30 $\pm$ 19.09} & \textcolor{gray}{367.61 $\pm$ 37.71} & \textcolor{gray}{19.65 $\pm$ 7.25} \\ 
\texttt{pole} & 700 & \textbf{958.58 $\pm$ 1.11} & \textcolor{gray}{207.46 $\pm$ 43.65} & \textcolor{gray}{589.01 $\pm$ 74.35} & \textcolor{gray}{123.30 $\pm$ 49.65} \\ 
\texttt{push} & 700 & \textbf{\textcolor{gray}{83.54 $\pm$ 164.92}} & \textcolor{gray}{-168.50 $\pm$ 45.46} & \textcolor{gray}{-144.62 $\pm$ 73.83} & \textcolor{gray}{-263.98 $\pm$ 54.44} \\ 
\texttt{cabinet} & 2500 & \textbf{\textcolor{gray}{664.13 $\pm$ 14.69}} & \textcolor{gray}{147.62 $\pm$ 34.00} & \textcolor{gray}{105.45 $\pm$ 52.28} & \textcolor{gray}{183.28 $\pm$ 63.76} \\ 
\texttt{highbar} & 750 & \textcolor{gray}{0.26 $\pm$ 0.05} & \textcolor{gray}{0.47 $\pm$ 0.19} & \textcolor{gray}{7.58 $\pm$ 2.11} & \textbf{\textcolor{gray}{18.43 $\pm$ 20.10}} \\ 
\texttt{door} & 600 & \textbf{\textcolor{gray}{270.92 $\pm$ 30.80}} & \textcolor{gray}{179.83 $\pm$ 91.33} & \textcolor{gray}{165.83 $\pm$ 104.86} & \textcolor{gray}{131.79 $\pm$ 12.89} \\ 
\texttt{truck} & 3000 & \textbf{\textcolor{gray}{1402.91 $\pm$ 49.02}} & \textcolor{gray}{1164.00 $\pm$ 38.29} & \textcolor{gray}{1341.04 $\pm$ 33.58} & \textcolor{gray}{1128.40 $\pm$ 16.26} \\ 
\texttt{cube} & 370 & \textcolor{gray}{33.79 $\pm$ 4.65} & \textcolor{gray}{104.14 $\pm$ 25.21} & \textcolor{gray}{63.57 $\pm$ 2.83} & \textbf{\textcolor{gray}{104.34 $\pm$ 32.21}} \\ 
\texttt{bookshelf\_simple} & 2000 & \textbf{\textcolor{gray}{805.64 $\pm$ 25.83}} & \textcolor{gray}{194.37 $\pm$ 33.46} & \textcolor{gray}{773.76 $\pm$ 33.82} & \textcolor{gray}{363.68 $\pm$ 71.15} \\ 
\texttt{bookshelf\_hard} & 2000 & \textbf{\textcolor{gray}{707.57 $\pm$ 45.21}} & \textcolor{gray}{64.19 $\pm$ 3.07} & \textcolor{gray}{577.13 $\pm$ 60.55} & \textcolor{gray}{300.05 $\pm$ 97.77} \\ 
\texttt{basketball} & 1200 & \textcolor{gray}{111.83 $\pm$ 126.20} & \textbf{\textcolor{gray}{120.66 $\pm$ 21.77}} & \textcolor{gray}{46.22 $\pm$ 26.74} & \textcolor{gray}{34.10 $\pm$ 10.00} \\ 
\texttt{window} & 650 & \textbf{714.22 $\pm$ 37.04} & \textcolor{gray}{63.27 $\pm$ 19.87} & \textcolor{gray}{158.37 $\pm$ 68.57} & \textcolor{gray}{65.68 $\pm$ 107.28} \\ 
\texttt{spoon} & 650 & \textbf{\textcolor{gray}{386.39 $\pm$ 0.64}} & \textcolor{gray}{70.69 $\pm$ 23.39} & \textcolor{gray}{331.83 $\pm$ 11.30} & \textcolor{gray}{118.28 $\pm$ 47.47} \\ 
\texttt{kitchen} & 4 & \textcolor{gray}{0.00 $\pm$ 0.00} & \textcolor{gray}{0.00 $\pm$ 0.00} & \textcolor{gray}{0.00 $\pm$ 0.00} & \textcolor{gray}{0.00 $\pm$ 0.00} \\ 
\texttt{package} & 1500 & \textcolor{gray}{-6957.24 $\pm$ 1154.91} & \textcolor{gray}{-7228.07 $\pm$ 467.74} & \textcolor{gray}{-8124.26 $\pm$ 337.79} & \textbf{\textcolor{gray}{-6946.94 $\pm$ 30.56}} \\ 
\texttt{powerlift} & 800 & \textbf{\textcolor{gray}{339.15 $\pm$ 0.22}} & \textcolor{gray}{92.48 $\pm$ 11.16} & \textcolor{gray}{260.93 $\pm$ 24.58} & \textcolor{gray}{128.70 $\pm$ 9.52} \\ 
\texttt{room} & 400 & \textbf{\textcolor{gray}{203.53 $\pm$ 13.99}} & \textcolor{gray}{84.20 $\pm$ 17.95} & \textcolor{gray}{123.92 $\pm$ 10.52} & \textcolor{gray}{14.50 $\pm$ 0.18} \\ 
\texttt{insert\_small} & 350 & \textbf{\textcolor{gray}{174.47 $\pm$ 4.34}} & \textcolor{gray}{117.24 $\pm$ 21.27} & \textcolor{gray}{115.93 $\pm$ 18.22} & \textcolor{gray}{11.87 $\pm$ 13.05} \\ 
\texttt{insert\_normal} & 350 & \textbf{\textcolor{gray}{214.55 $\pm$ 4.12}} & \textcolor{gray}{176.66 $\pm$ 10.58} & \textcolor{gray}{144.09 $\pm$ 18.02} & \textcolor{gray}{47.00 $\pm$ 67.78} \\ \bottomrule
\end{tabular}
\end{table}

\end{document}